\documentclass[conf]{new-aiaa}
\usepackage[utf8]{inputenc}

\usepackage[table,xcdraw]{xcolor}
\usepackage{graphicx}
\usepackage{amsmath}
\usepackage[version=4]{mhchem}
\usepackage{siunitx}
\usepackage{longtable,tabularx}
\usepackage{subcaption}
\usepackage[flushleft]{threeparttable}
\usepackage{multirow}
\usepackage[framed,autolinebreaks,useliterate]{mcode}
\usepackage[absolute,overlay]{textpos}

\title{UAS Simulator for Modeling, Analysis and Control\\in Free Flight and Physical Interaction}

\author{Azarakhsh Keipour\footnote{Applied Scientist, Amazon Robotics, keipour@gmail.com. The publication was written prior to A. Keipour joining Amazon. During the realization of this work, he was affiliated with the Robotics Institute at Carnegie Mellon University.}}
\affil{Amazon, Arlington, VA 22202}
\author{Mohammadreza Mousaei\footnote{Ph.D. Student, The Robotics Institute, mmousaei@andrew.cmu.edu.} and Dongwei Bai\footnote{Master's Student, Department of Mechanical Engineering, dongweib@andrew.cmu.edu.} and Junyi Geng\footnote{Postdoctoral Fellow, The Robotics Institute, junyigen@andrew.cmu.edu.} and Sebastian Scherer\footnote{Associate Research Professor, The Robotics Institute, basti@cmu.edu.}}
\affil{Carnegie Mellon University, Pittsburgh, PA 15213}

\begin{document}

\begin{textblock*}{15cm}(4.75cm,1cm) % {block width} (coords) 
   {\Large\textcolor{red}{AIAA SciTech Forum, January 23-27 2023}\\
   \textcolor{purple}{\url{https://arc.aiaa.org/doi/10.2514/6.2023-1279}}}
\end{textblock*}

\maketitle

\begin{abstract}
This paper presents the ARCAD simulator for the rapid development of Unmanned Aerial Systems (UAS), including underactuated and fully-actuated multirotors, fixed-wing aircraft, and Vertical Take-Off and Landing (VTOL) hybrid vehicles. The simulator is designed to accelerate these aircraft's modeling and control design. It provides various analyses of the design and operation, such as wrench-set computation, controller response, and flight optimization. In addition to simulating free flight, it can simulate the physical interaction of the aircraft with its environment. The simulator is written in MATLAB to allow rapid prototyping and is capable of generating graphical visualization of the aircraft and the environment in addition to generating the desired plots. It has been used to develop several real-world multirotor and VTOL applications. The source code is available at \url{https://github.com/keipour/aircraft-simulator-matlab}.
\end{abstract}

% \section{Nomenclature}

% {\renewcommand\arraystretch{1.0}
% \noindent\begin{longtable*}{@{}l @{\quad=\quad} l@{}}
% $A$  & amplitude of oscillation \\
% $a$ &    cylinder diameter \\
% $C_p$& pressure coefficient \\
% $Cx$ & force coefficient in the \textit{x} direction \\
% $Cy$ & force coefficient in the \textit{y} direction \\
% c   & chord \\
% d$t$ & time step \\
% $Fx$ & $X$ component of the resultant pressure force acting on the vehicle \\
% $Fy$ & $Y$ component of the resultant pressure force acting on the vehicle \\
% $f, g$   & generic functions \\
% $h$  & height \\
% $i$  & time index during navigation \\
% $j$  & waypoint index \\
% $K$  & trailing-edge (TE) nondimensional angular deflection rate
% \end{longtable*}}

\section{Introduction} \label{sec:introduction}

\lettrine{T}{he} past few decades have seen rapid growth in Unmanned Aerial Systems (UAS) research. New applications are introduced using these systems' autonomy and flying capabilities to improve the existing processes, reduce the costs, or perform tasks that were not possible before~\cite{Balaram2021, en15010217, Keipour:2022:sensors:mbzirc, RAKHA2018252}. New aircraft designs have been developed, including fully-actuated multirotors and hybrid VTOLs, while applications have extended into physical interaction of the aircraft with their environment for tasks ranging from non-destructive contacts to manipulation~\cite{Park2018a, Keipour:2022:thesis, Rashad2019a, Trujillo2019, Mousaei:2022:iros:vtol, Ollero2018, Keipour:2020:arxiv:integration}.

In response to the advances in the UAS designs and applications, many new simulators have been developed for various purposes~\cite{song2020flightmare, li2022rotortm}. The most common simulator is Gazebo~\cite{2012simpar_meyer}. It is a general robotics simulator that has been repurposed for simulating UASs and provides all the general functionalities expected from a simulator. However, it is difficult to use for testing new aircraft architectures and lower-level control designs and is time-consuming to extend. Many extensions have been developed to expand Gazebo's use for aircraft (e.g., RotorS~\cite{Furrer2016}). While Gazebo can be a perfect simulator for many tasks, it is unsuitable for prototyping new designs, controllers, and interactive applications.

Another common simulator is AirSim~\cite{airsim2017fsr} which only focuses on visually realistic environments and is mainly used for vision-oriented applications and high-level trajectory tests. On the other side of the spectrum are simulators that focus on accurate dynamics for larger aircraft and cannot be used for smaller UAS aircraft such as fully-actuated multirotors (e.g., FlightGear~\cite{perry2004flightgear} and XPlane~\cite{x_2022}). All these simulators are difficult to use for prototyping, and making any changes is time-consuming and requires tedious work.

We developed a complete UAS simulator that accelerates the prototyping of new aircraft architectures and controller designs and allows rapid testing of new applications, including free flight or physical interaction with the environment. The ARCAD (AirLab Rapid Controller and Aircraft Design) simulator is written in MATLAB and is fully modular to facilitate prototyping. It can provide various analyses of the aircraft design and its operation, such as wrench set computation, step responses, and force/torque feedback to optimize different aspects of the architecture, control system, and application. In addition to allowing the plotting of all the internal and external signals, the simulator can provide a graphical visualization of the experiments. We have designed new fully-actuated vehicles and hybrid VTOLs with our simulator and have used it for developing real-world applications to interact with the physical world and for aircraft failure recovery~\cite{Keipour:2022:thesis, Keipour:2020:arxiv:integration, Mousaei:2022:iros:vtol}.

\section{Features} \label{sec:features}

The ARCAD simulator has a hierarchical modular design to facilitate testing new aircraft models and controllers. For example, the controller module can be seamlessly replaced by another controller, just keeping the same interface. Inside the controller module, each sub-module (e.g., attitude controller or position controller) can also be independently switched by a new sub-module.

This section briefly describes some of the capabilities of the simulator. The simulator can be further extended to support new types of vehicles, controllers, and analyses. For the interested reader, a workshop paper~\citep{Keipour:2023:icra-workshop:simulator} has an overview of the internal workings of the simulator. 

\subsection{Aircraft Design}

\begin{figure}[!htb]
\centering
    \begin{subfigure}[b]{0.24\textwidth}
        \includegraphics[width=\textwidth]{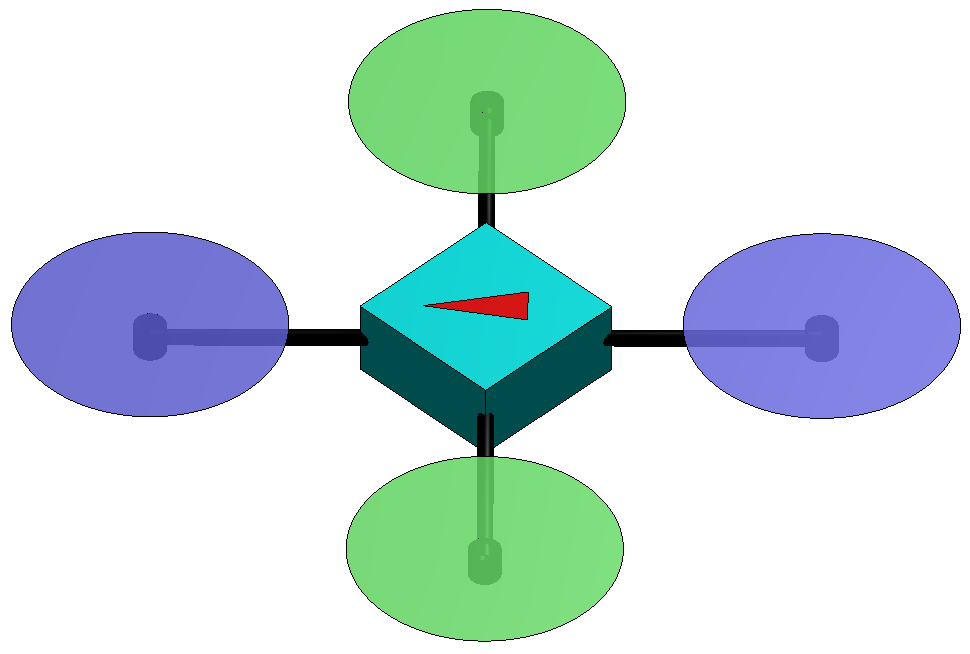}
        \caption{~}
    \end{subfigure}
    \hfill
    \begin{subfigure}[b]{0.24\textwidth}
        \includegraphics[width=\textwidth]{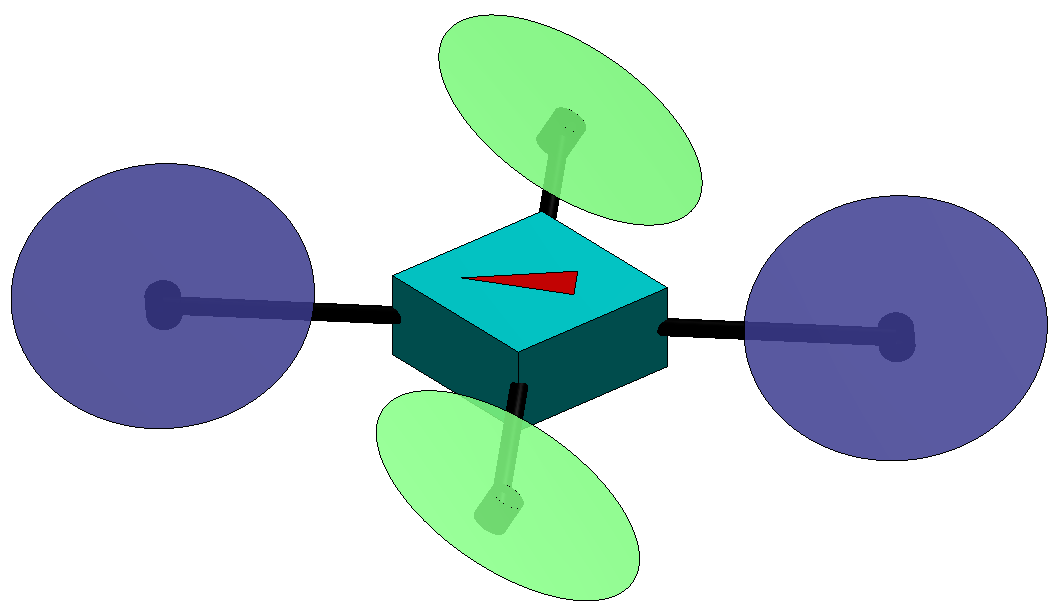}
        \caption{~}
    \end{subfigure}
    \hfill
    \begin{subfigure}[b]{0.24\textwidth}
        \includegraphics[width=\textwidth]{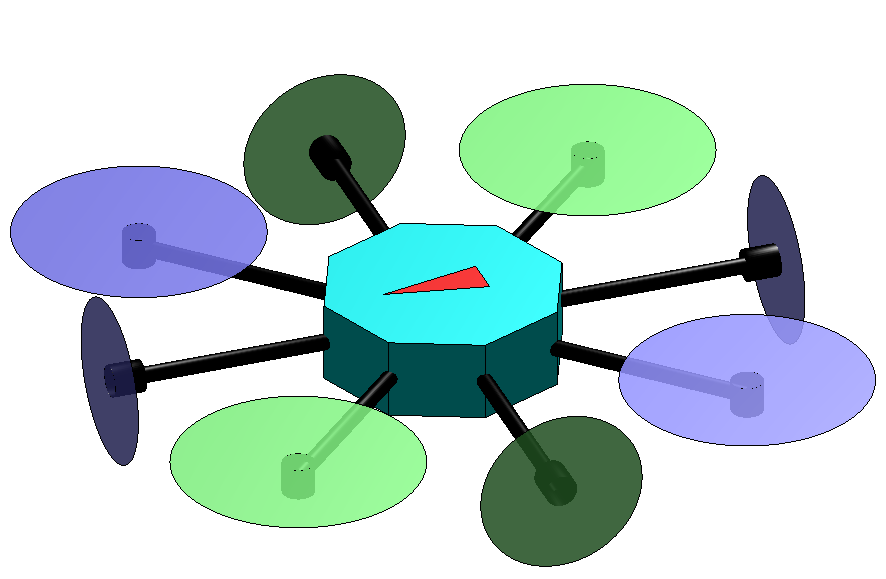}
        \caption{~}
    \end{subfigure}
    \hfill
    \begin{subfigure}[b]{0.24\textwidth}
        \includegraphics[width=\textwidth]{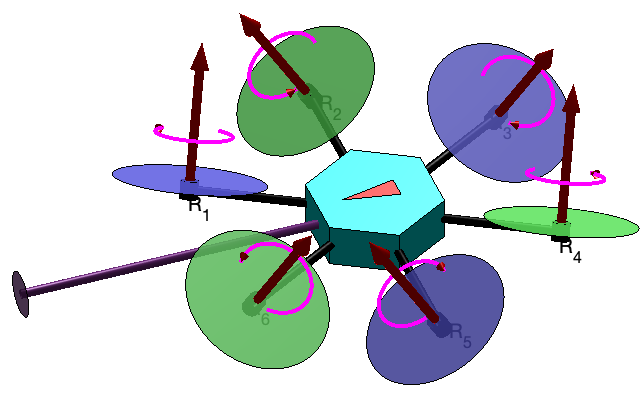}
        \caption{~}
        \label{fig:designs-fa-hex}
    \end{subfigure}
    
    \medskip
    \begin{subfigure}[b]{0.32\textwidth}
        \includegraphics[width=\textwidth]{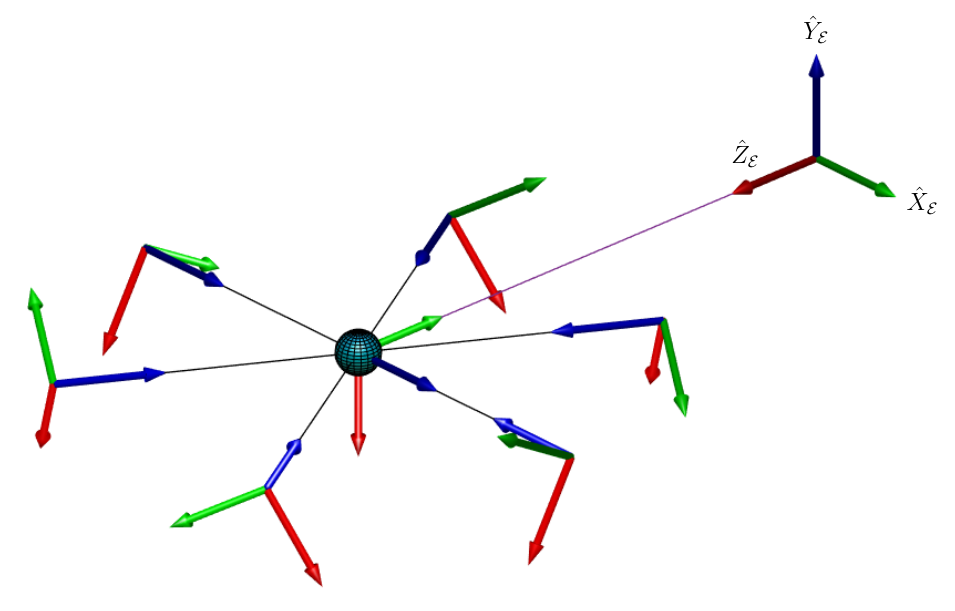}
        \caption{~}
    \end{subfigure}
    \hfill
    \begin{subfigure}[b]{0.32\textwidth}
        \includegraphics[width=\textwidth]{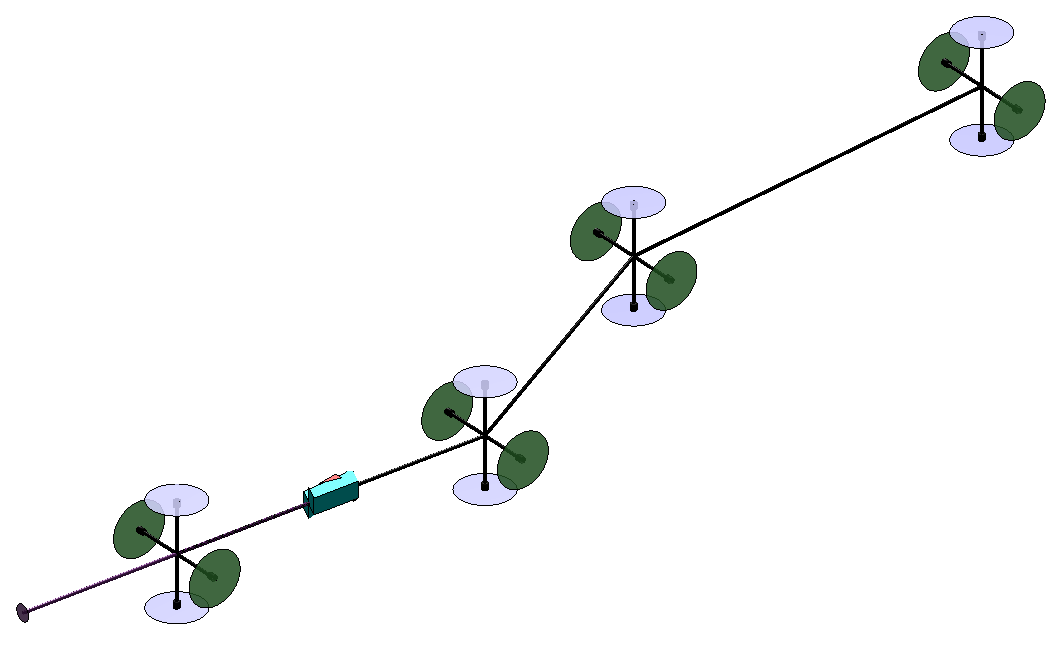}
        \caption{~}
    \end{subfigure}
    \hfill
    \begin{subfigure}[b]{0.32\textwidth}
        \includegraphics[width=\textwidth]{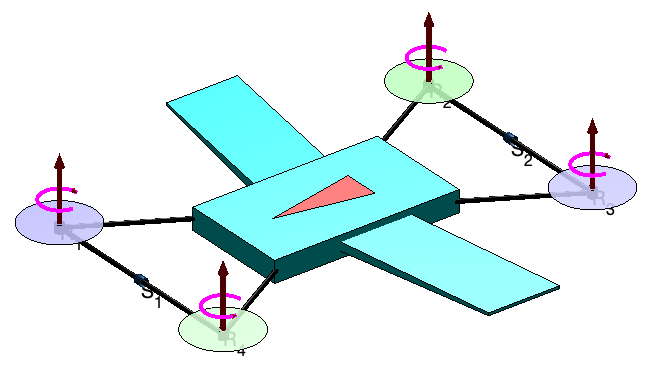}
        \caption{~}
        \label{fig:designs-vtol}
    \end{subfigure}
\caption{Visualization of some aircraft architectures defined in our simulator. (a) A quadrotor. (b) A quadrotor with variable-pitch rotors (i.e., thrusters). (c) A fully-actuated octorotor. (d) A fully-actuated hexarotor with tilted arms and a manipulation arm, visualized with its rotor axes and rotation directions. (e) Visualization of only the axes for the hexarotors in (d). (f) A multilink multirotor aircraft. (e) A hybrid VTOL with variable pitch rotors.}
\label{fig:designs}
\end{figure}

Our simulator supports any arrangement of fixed and variable-pitch rotors for multirotors provided by the user with minimal coding. It then automatically generates the geometry and control allocation matrix for the given multirotor description. Similarly, fixed-wing and hybrid VTOL aircraft can be defined with minimal descriptions. The default aircraft dynamic and aerodynamic models can easily be modified or extended in the simulator. The example code below defines the fully-actuated hexarotor of Figure~\ref{fig:designs-fa-hex} that has symmetric rotor arm placements with $30^\circ$ tilted rotor arms, a manipulator arm, and no dihedral or inward rotor angles.

\begin{lstlisting}
RotorPlacementAngles = [30, 90, 150, 210, 270, 330];
RotorRotationDirections = [-1, 1, -1, 1, -1, 1];
RotorSidewardAngle = [-30, 30, -30, 30, -30, 30];
m = multirotor(RotorPlacementAngles, RotorRotationDirections);
m.SetRotorAngles(0, RotorSidewardAngle, 0);
m.AddEndEffector(arm);
\end{lstlisting}

Figure~\ref{fig:designs} illustrates some aircraft designs defined in the simulator.

\subsection{Controller Design}

\begin{figure}[!htb]
\centering
    \begin{subfigure}[b]{0.48\textwidth}
        \includegraphics[width=\textwidth]{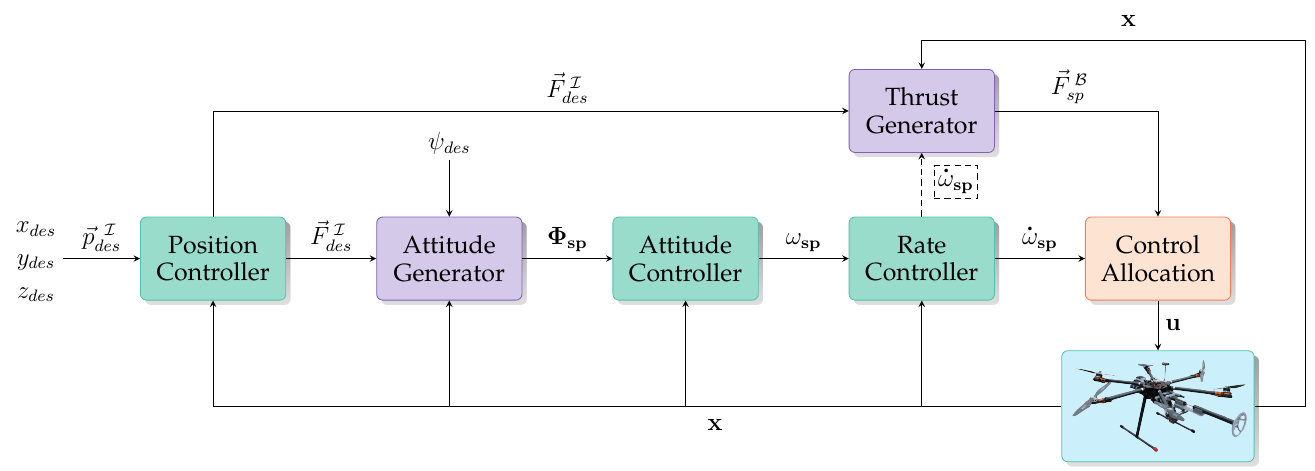}
        \caption{~}
    \end{subfigure}
    \hfill
    \begin{subfigure}[b]{0.48\textwidth}
        \includegraphics[width=\textwidth]{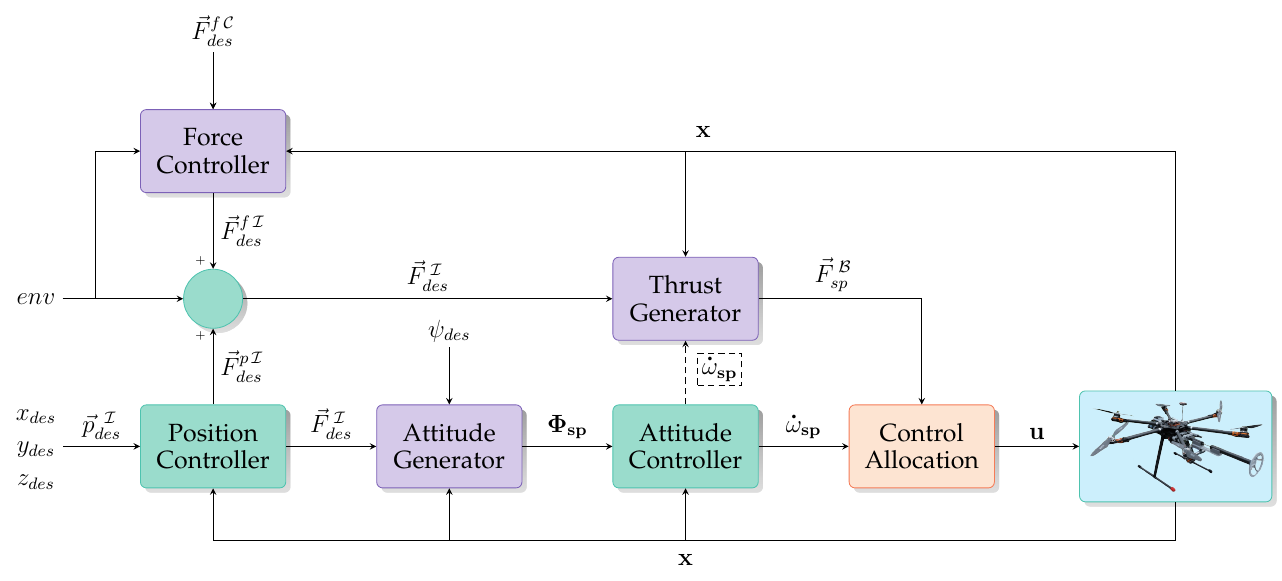}
        \caption{~}
    \end{subfigure}
\caption{The controller architectures for multirotors. (a) Free-flight controller. (b) Hybrid Force-Position controller for physical interaction applications.}
\label{fig:controllers}
\end{figure}

The modular nature of the simulator facilitates the development and testing of new controllers. Some default controllers are provided for multirotors, fixed-wings, and hybrid VTOLs that work with a wide range of new aircraft designs and various applications. The default controller module includes sub-modules for attitude controller, position controller, control allocation, and force controller. Figure~\ref{fig:controllers} illustrates the default controller architectures for multirotors in free-flight and physical interaction applications. 

In addition to these flight controllers, a trajectory controller is provided to control trajectories comprised of the desired aircraft poses, servo angles (for variable-pitch rotors), and wrenches (forces and moments) applied to the environment during the physical contact. Moreover, the complete set of attitude strategies (see~\cite{Keipour:2020:arxiv:integration}) is provided for fully-actuated multirotors, and an MPC-based controller is implemented for the fixed-wing aircraft.

\begin{figure}[!htb]
\centering
\includegraphics[width=0.7\linewidth]{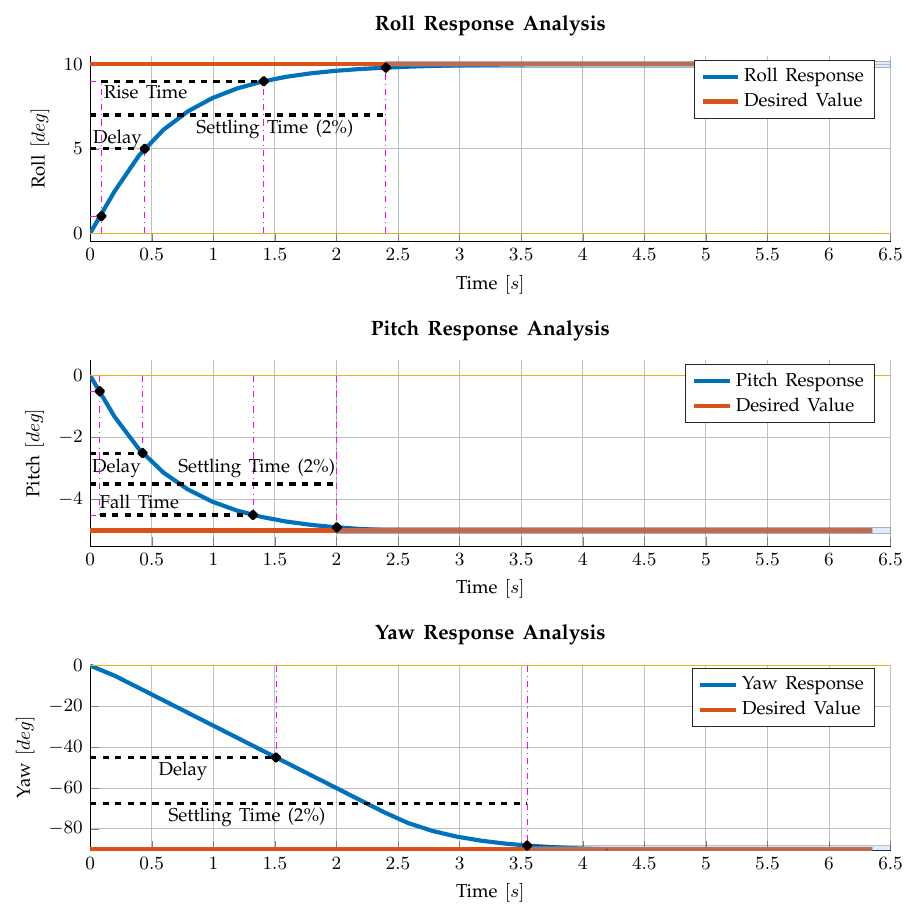}
\caption{Automatic analysis of the attitude response for the hexarotor with tilted arms.}
\label{fig:attitude-response}
\end{figure}

Several tools are provided to assist with controller design and tuning that automatically analyze and visualize the controller module. Figure~\ref{fig:attitude-response} illustrates a sample attitude response analysis provided by our simulator. The plots show the response to the desired input of $10^\circ$ roll, $-5^\circ$ pitch, and $-90^\circ$ yaw given to the fully-actuated hexarotor of Figure~\ref{fig:designs-fa-hex}. The system automatically provides similar plots for different controller modules and analyzes the response metrics on the plot and as the output to allow manual or automated tuning of the controller.

\subsection{Analysis}

\begin{figure}[!htb]
\centering
\includegraphics[width=0.9\linewidth]{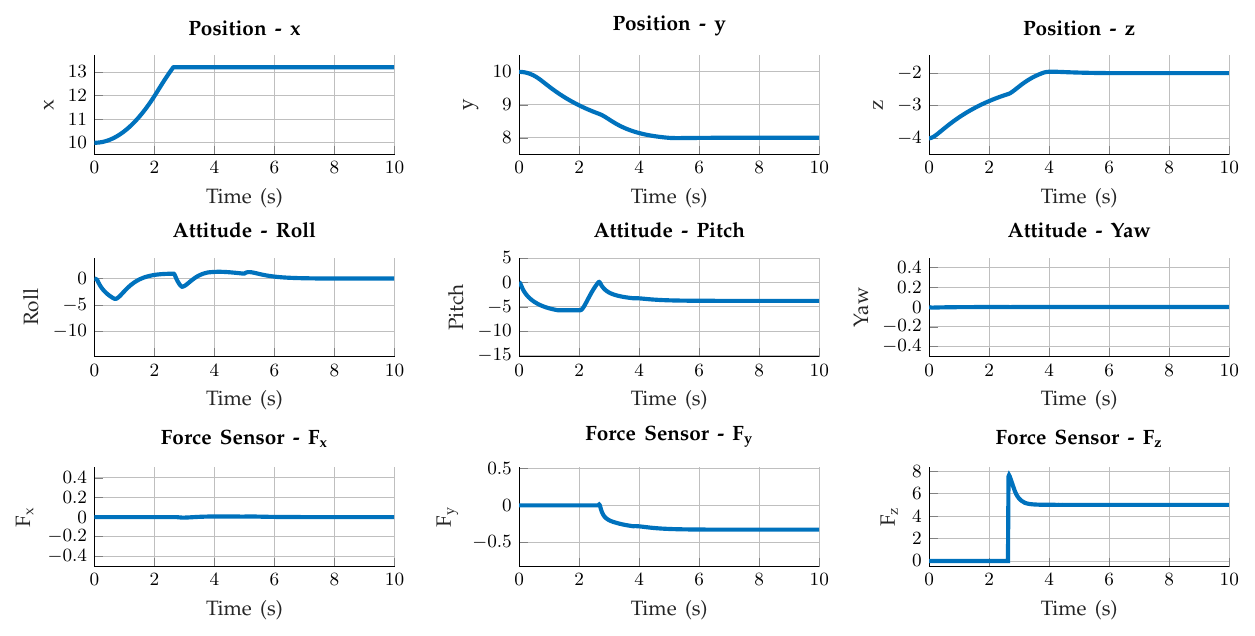}
\caption{A sample plot generated for a multirotor applying a 5~N force to a wall.}
\label{fig:sample-plots}
\end{figure}

For each simulation scenario, all the state and internal signals are available for plotting and analysis, including desired and measured aircraft and end-effector states and generated and interaction wrenches. A single-line command can describe the desired signals to plot. Figure~\ref{fig:sample-plots} shows the desired pose and force signals asked by the user with a single-line command. The plots are the measured position in inertial frame axes, the attitude of the robot (roll, pitch, and yaw), and the forces measured at the end-effector in the end-effector frame for the robot of Figure~\ref{fig:designs-fa-hex} trying to apply a 5~N force to a wall.

\begin{figure}[!htb]
\centering
    \begin{subfigure}[b]{0.30\textwidth}
        \includegraphics[width=\textwidth]{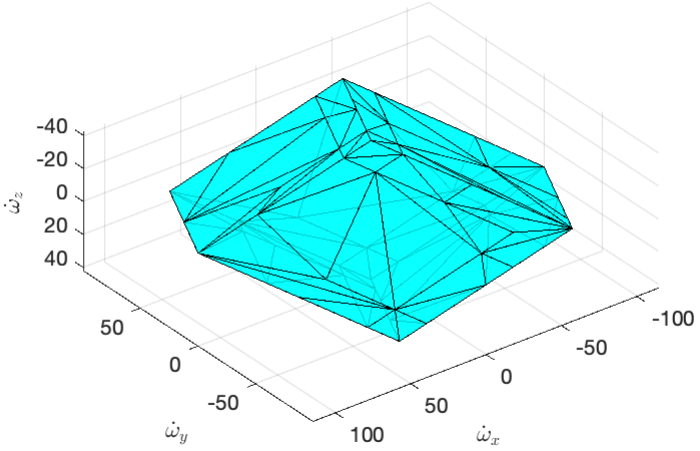}
        \caption{~}
    \end{subfigure}
    ~
    \begin{subfigure}[b]{0.155\textwidth}
        \includegraphics[width=\textwidth]{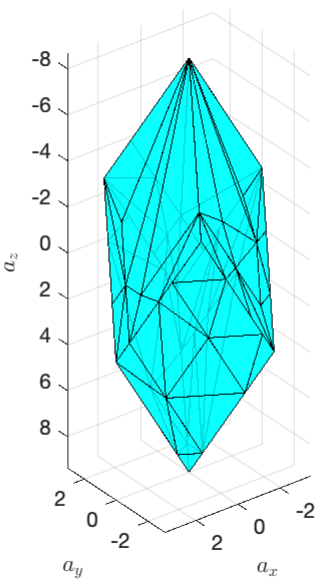}
        \caption{~}
    \end{subfigure}
    ~
    \begin{subfigure}[b]{0.35\textwidth}
        \includegraphics[width=\textwidth]{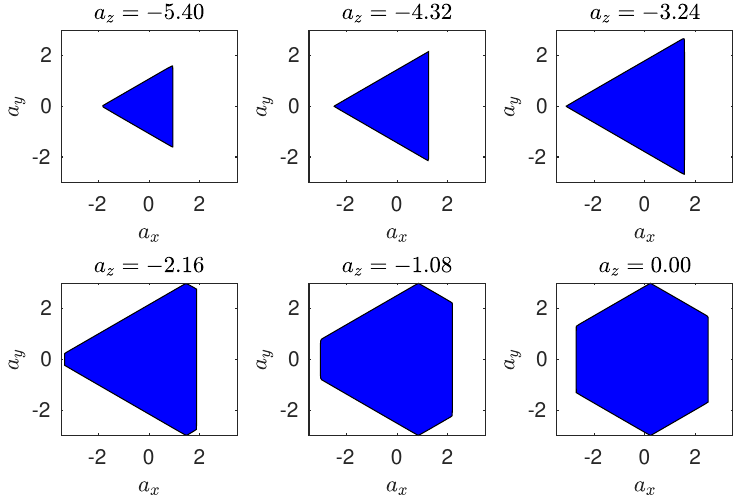}
        \caption{~}
    \end{subfigure}
    
    \medskip
    
    \begin{subfigure}[b]{0.30\textwidth}
        \includegraphics[width=\textwidth]{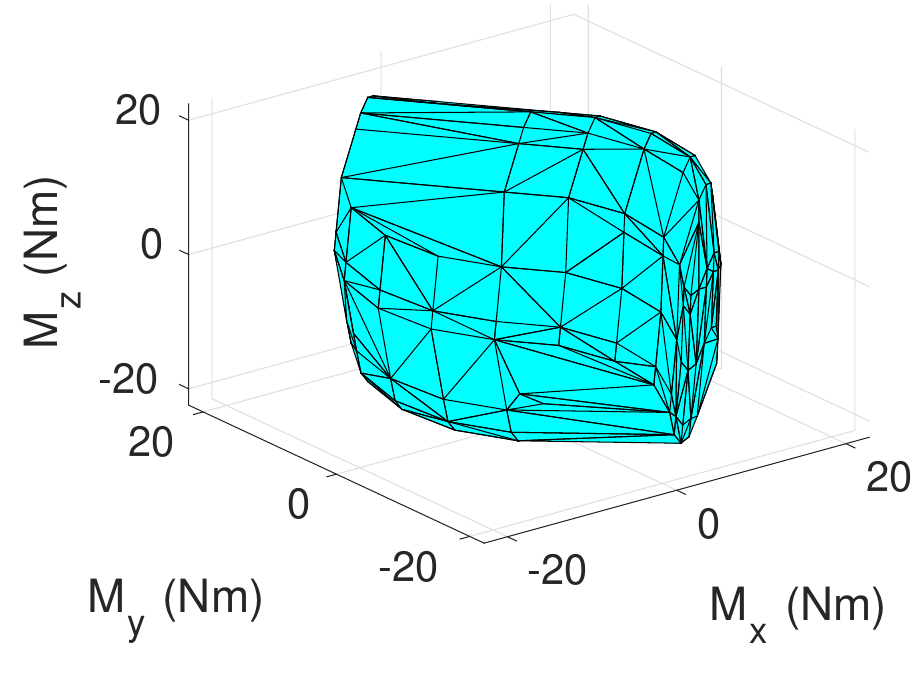}
        \caption{~}
    \end{subfigure}
    ~
    \begin{subfigure}[b]{0.155\textwidth}
        \includegraphics[width=\textwidth]{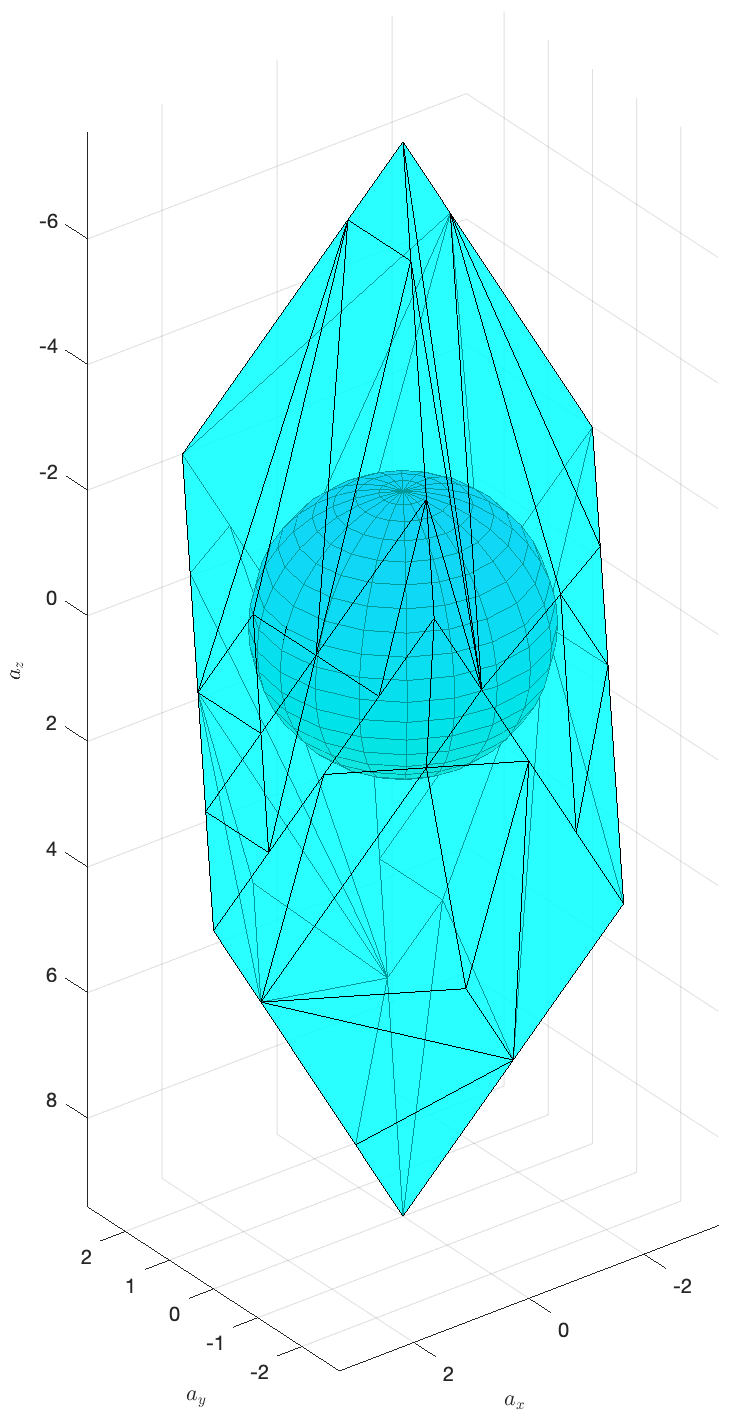}
        \caption{~}
    \end{subfigure}
    ~
    \begin{subfigure}[b]{0.30\textwidth}
        \includegraphics[width=\textwidth]{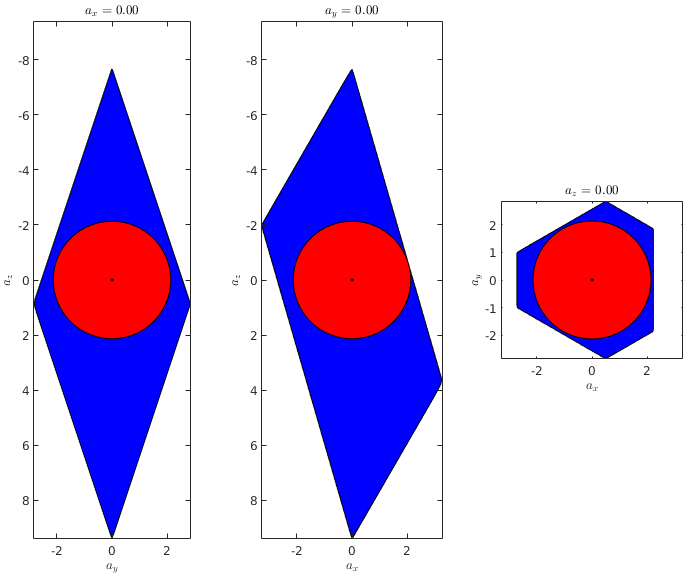}
        \caption{~}
    \end{subfigure}
    \caption{Wrench set and omni-directional acceleration analysis for aircraft. (a) Moment set for the hexarotor with tilted arms. (b, c) Force set and cross-sections (along $Z$ axis) for the hexarotor with tilted arms. (d) Moment set for the hybrid VTOL. (e, f) Omni-directional acceleration sphere and cross-sections (along all three axes) for the hexarotor with tilted arms.}
\label{fig:analysis}
\end{figure}

Furthermore, the simulator can analyze the wrench set (i.e., force and moment polytopes) in real-time~\cite{Keipour:2023:unpub:wrench}. This tool allows aircraft design optimization and provides several benefits, including planning for physical interaction, failure recovery, and flight optimization (see~\cite{Keipour:2022:thesis} for descriptions of applications). In addition, the simulator computes metrics such as omni-directional accelerations and available lateral forces for fully-actuated multirotors. Figure~\ref{fig:analysis} shows sample wrench sets and analysis computed and visualized by the simulator. The plots illustrate the moment sets for the fully-actuated hexarotor of Figure~\ref{fig:designs-fa-hex} and the hybrid VTOL of Figure~\ref{fig:designs-vtol}. They also show the force set with its cross-sections along the $Z$ axis and the omni-directional acceleration sphere with its cross-sections along all three axes for the hexarotor of Figure~\ref{fig:designs-fa-hex}.

\subsection{Applications}

The ARCAD simulator was initially developed for modeling, control design, and analysis. However, over time we have extended it to support complete experiments to improve our applications before implementing them on real aircraft. The extensions enable environment design (e.g., adding 3-D obstacles with collision definition), add visual capabilities (e.g., 3-D and POV camera views, textures), allow monitoring of signals at each moment (e.g., artificial horizon and on GUIs), provide higher-level controllers for trajectories and tasks, and allow recording of the outputs (e.g., data and high-quality videos of the scenario). 

We have used the simulator for simpler applications such as trajectory following among obstacles to more advanced applications such as force-controlled writing on the wall using a fully-actuated multirotor~\cite{Keipour:2022:thesis}, real-time wrench-set analysis~\cite{Keipour:2022:thesis, Keipour:2023:unpub:wrench, Keipour:2022:icra-workshop:doo}, Model Predictive Path Integral (MPPI) control during physical interaction~\cite{Mousaei:2023:unpub:mppi} and real-time failure recovery for hybrid VTOL aircraft~\cite{Mousaei:2022:iros:vtol, Mousaei:2022:icra-workshop:vtol}. 

\begin{figure}[!htb]
\centering
\includegraphics[width=\linewidth]{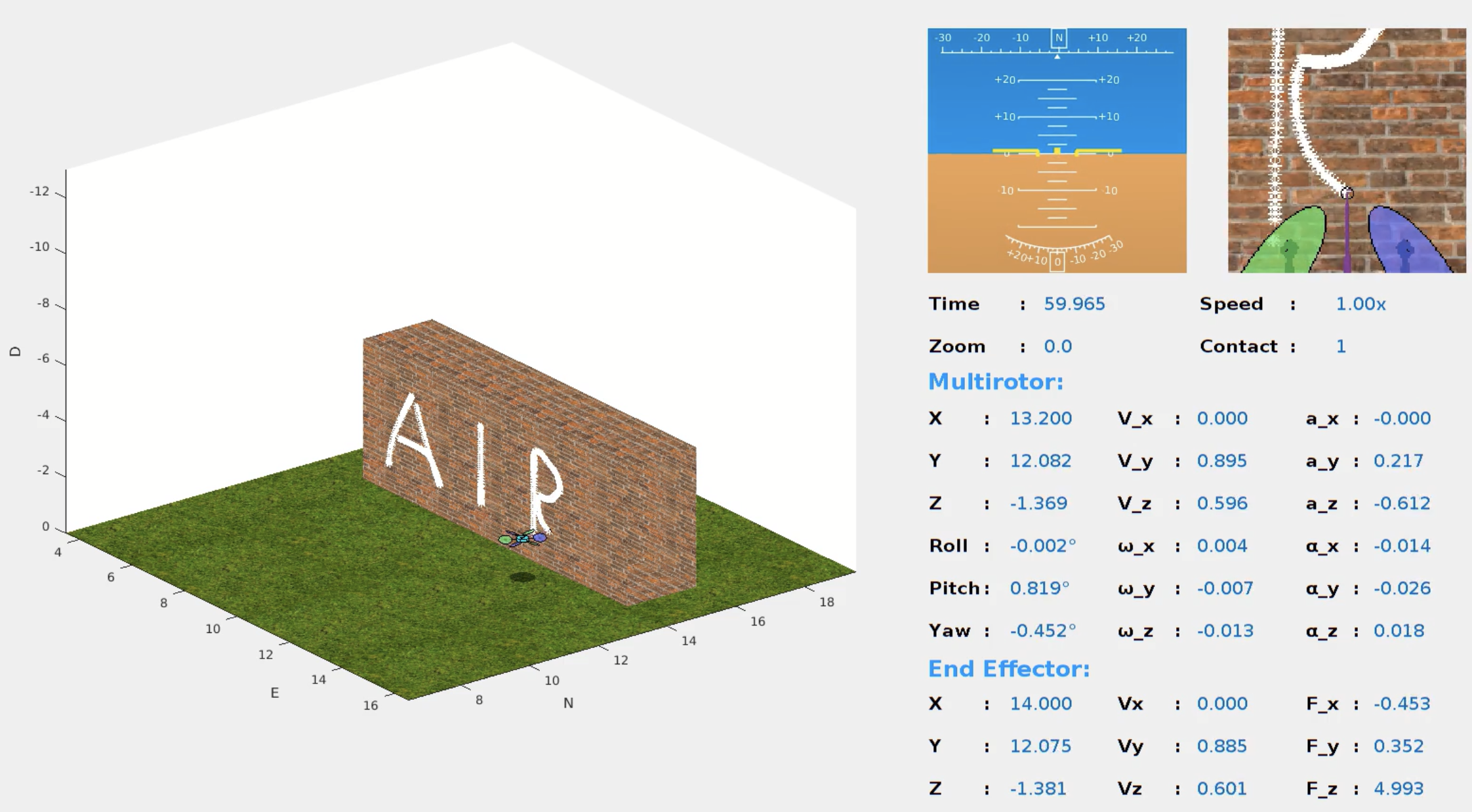}
\caption{A multirotor writing letters on the wall while controlling the applied force to be exactly 5~N.}
\label{fig:applications}
\end{figure}

Figure~\ref{fig:applications} shows a screenshot of our simulator running experiments. The snapshot shows the hexarotor of Figure~\ref{fig:designs-fa-hex} writing AIR on the wall with the desired force of $5$~N. 

\section{Conclusion} \label{sec:conclusion}

This paper describes our simulator for rapid aircraft design and development. We explained the features useful for aircraft and controller design and went over some of the analyses provided by the simulator. The simulator has been used to develop novel fully-actuated multirotors and hybrid VTOL aircraft in our team. It has assisted us with developing novel controllers and real-world applications, such as writing on the wall, failure recovery, and contact inspection. The source code for the simulator is provided with this publication and can be accessed from \url{https://github.com/keipour/aircraft-simulator-matlab}.

In future iterations, there are plans to allow communication through ROS and MAVLINK. Additionally, we plan to provide more pre-written controllers and analysis to allow rapid comparison of different controller methods. We believe this simulator can accelerate the research and development of new UASs and control methods and reduce the costs of implementing new applications.

\section*{Acknowledgments}
This work was partially supported through NASA Grant Number 80NSSC19C010401 and sponsored by Carnegie Mellon University Robotics Institute.
\bibliography{references}

\end{document}